# Compression, Restoration, Resampling, "Compressive Sensing": Fast Transforms in Digital Imaging

L. P. Yaroslavsky,
Dept. of Physical Electronics, School of Electrical Engineering,
Tel Aviv University, Tel Aviv 69978, Israel

**Abstract**

Transform image processing methods are methods that work in domains of image transforms, such as Discrete Fourier, Discrete Cosine, Wavelet and alike. They are the basic tool in image compression, in image restoration, in image resampling and geometrical transformations and can be traced back to early 1970-ths. The paper presents a review of these methods with emphasis on their comparison and relationships, from the very first steps of transform image compression methods to adaptive and local adaptive transform domain filters for image restoration, to methods of precise image resampling and image reconstruction from sparse samples and up to "compressive sensing" approach that has gained popularity in last few years. The review has a tutorial character and purpose.

# 1. Introduction

## 1.1. A bit of a historical perspective

It has passed roughly half a century since first attempts to use recently emerged digital computers for image processing. Nowadays imaging engineering has overwhelmingly become digital and computational.

It will not be an exaggeration to assert that digital image processing came into being with introduction about 50 years of the **Fast Fourier Transform** algorithm (FFT, [ 1]) for computing the **Discrete Fourier Transform** (DFT). This publication immediately resulted in impetuous growth of activity in all branches of digital signal and image processing and their applications.

The second wave in this process was inspired by the introducing into communication and computer engineering, in 1970s, of **Walsh-Hadamard transform** and **Haar transform** ([ 2]) and the development of a large family of fast transforms with FFT-type algorithms ([ 3],[ 4],[5]). Whereas Walsh-Hadamard and Haar transforms had already been known in mathematics, other transforms, such as, for instance, quite popular at the time **Slant Transform** ([ 6]), were being invented "from scratch". This development was mainly driven by the needs of image data compression, though the usefulness of transform domain processing for image restoration and enhancement was also recognized very soon ([ 3]. This period ended up with the introduction of the **Discrete Cosine Transform** (DCT, **[** 7], [ 8] ), which was soon widely recognized as the best choice among all available at the time transforms, and resulted in JPEG and MPEG standards for image, audio and video compression.

The next milestone in transform signal processing was the introduction, in the 1980-th, of a large family of new transforms that are known, due to J. P. Morlet, as wavelets ([9]). A rapid burst of works followed publications by J. P. Morlet, A. Grossman, Y. Meyer, I. Daubechies, S. Mallat. This development continued the line of inventing new transforms better suited for the purposes of signal and image processing. Specifically, the main motivation was to achieve a better local representation of signals in contrast to the "global" representation that is characteristic to Fourier, DCT and Walsh-Hadamard Transforms.

A common method in designing new transform is generating basis functions of the transform from a primary, or "mother" function by means of its certain modifications. The simplest way for such a modification is the coordinate shift. This is how, for instance, sampling basis functions are formed. Yet another possible simple method is coordinate scaling. The above-mentioned fast transforms, with one exception, implement just this coordinate scaling method. The exception is Haar transform. Haar transform is built upon combining these two methods, coordinate shifting and scaling ones. This combination is the method that gave rise to wavelets and imparted them their most attractive feature, that of multi-resolution.

Since their introduction, wavelets have gained a great popularity. During 1980-1990th a vast variety of discrete wavelet transforms was suggested ([ 10]) for solving various tasks in signal and in image processing. In late 90's there were even claims that wavelets have made the Fourier Transform obsolete. Of course, this was an over-exaggeration. Undoubtedly, however, that nowadays wavelets, alongside other above mentioned transforms, constitute a well-established tool kit that has a wide range of applications in digital imaging and image processing.

In this review we expose fast transform methods for solving primary digital imaging tasks: image compression, restoration and resampling.

### *1.2. Why transforms? Which transforms?*

The main distinctive feature of transforms that makes them so efficient in digital imaging is their ***energy compaction capability***. In regular image representations in form of sets of ordered pixels, some pixels, for instance, those that belong to object borders, are more important than others and there are always some number of pixels in each particular image that are of no importance, can be dropped out from image representation and restored from the remaining "important" pixels. But the problem is that one never knows in advance which pixels in the image are "important" and which are not.

The situation is totally different in image representation in transform domain. For orthogonal transforms that feature good energy compaction capability, a lion share of total image "energy" (sum of squared transform coefficients) is concentrated in a small fraction of transform coefficients, which indices are, as a rule, known in advance for the given type of images or can be easily detected. It is this feature of transforms that is called their ***energy compaction capability***. It allows replacing images with their "band-limited", in terms of a specific transform, approximations, i.e. approximations defined by a sufficiently small fraction of image transform coefficients.

Customarily, the "band-limited" image approximation accuracy is evaluated in terms of the root mean square approximation error (RMSE) for images that are subjects of processing, or ***image ensembles***. By virtue of this, the optimal transform that secures the least band limited approximation error is defined by the ***ensemble correlation function***, i.e. image autocorrelation function averaged over the image ensemble. For continuous (not sampled signals), this transform is called the ***Karhunen-Loeve Transform*** ([ 11],[ 12]). Its discrete analog for sampled signals is called the ***Hotelling transform*** ([ 13]) and the result of applying this transform to sampled images is called signal ***principal component decomposition***. Karhunen-Loeve and Hotelling transforms provide the theoretical lower bound for compact (in terms of the number of components in signal decomposition) signal discrete representation for the RMSE criterion of image approximation.

However, being optimal in terms of the energy compaction capability, Karhunen-Loeve and Hotelling transforms have, generally, high computational complexity: the per pixel number of operations required for their computation is of the order of the image size. This why for practical needs only fast transforms that feature so called "fast transform" algorithms with computational complexity of the order of logarithm of the image size or lower are considered. A register of the most relevant fast transforms is presented in Table 1. Their mathematical definitions are given in Appendix.

Table 1. Fast transforms, their main characteristic features and application areas

|  | Relevance to imaging optics | Main characteristic features | Main application areas in digital imaging |
|---|---|---|---|
| Discrete Fourier Transforms (DFT) | Represent optical integral Fourier Transform | Cyclic shift invariance Vulnerable to boundary effects | - Analysis of periodicities<br>- Fast convolution and correlation<br>- Fast and accurate image resampling<br>- Image compression<br>- Image denoising and deblurring<br>- Numerical synthesis and reconstruction of holograms in digital holography<br>- Image reconstruction from projections in computed tomography |
| Discrete Cosine Transform (DCT) | Represents optical integral Fourier Transform | Cyclic shift invariance (with a double cycle); Virtually not sensitive to boundary effects Good energy compaction capability | |
| Walsh-Hadamard Transform | No direct relevance | Binary basis functions. Provides piece-wise constant separable image band-limited approximation. | -Image compression (marginal)<br>-Signal and image denoising (marginal)<br>-Coded aperture imaging |
| Discrete Wavelet Transforms | Signal sub-band decomposition | Multi-resolution. Good energy compaction capability | - Image compression<br>- Signal/image wide band noise denoising<br>- Image subsampling |

Different fast transforms have different energy compaction capability for different types of images. Figure 1 illustrates the energy compaction capabilities of Discrete Fourier, Discrete Cosine, Walsh and Haar transforms on a particular test image. The figure shows that for this particular image DCT demonstrates the best energy compaction capability: 95% of the image energy is contained in only 7.4% of all DCT spectral coefficients, whereas for DFT, Walsh and Haar transforms these fractions are 9.6%, 14% and 8.1%, correspondingly.

Experience shows that DCT is, for majority of images, one of transforms with the highest energy compaction capability. The competing transforms are appropriately designed wavelets ([ 10]). This property of DCT has a simple and intuitive explanation. DCT of a discrete signal is essentially, to the accuracy of an unimportant exponential phase factor, DFT of the same signal extended outside its borders to a double length by means of its inversed, in the order of samples, copy ([ 14], [ 15], [ 18]). DFT implies periodical extension of signals, which may cause severe discontinuities at signal borders and thus may require excessive high frequency components to reproduce them. DCT also implies periodical extension, however of not the signal itself but of its above described extended copy. Thanks to such an "even" extension, the extended signal has no discontinuities at its borders, as well as at borders of the initial signal. Therefore such an extended signal is "smooth" at the borders and its DFT spectrum, i.e. DCT spectrum of the initial signal, decays to zero faster than DFT spectrum of this signal.

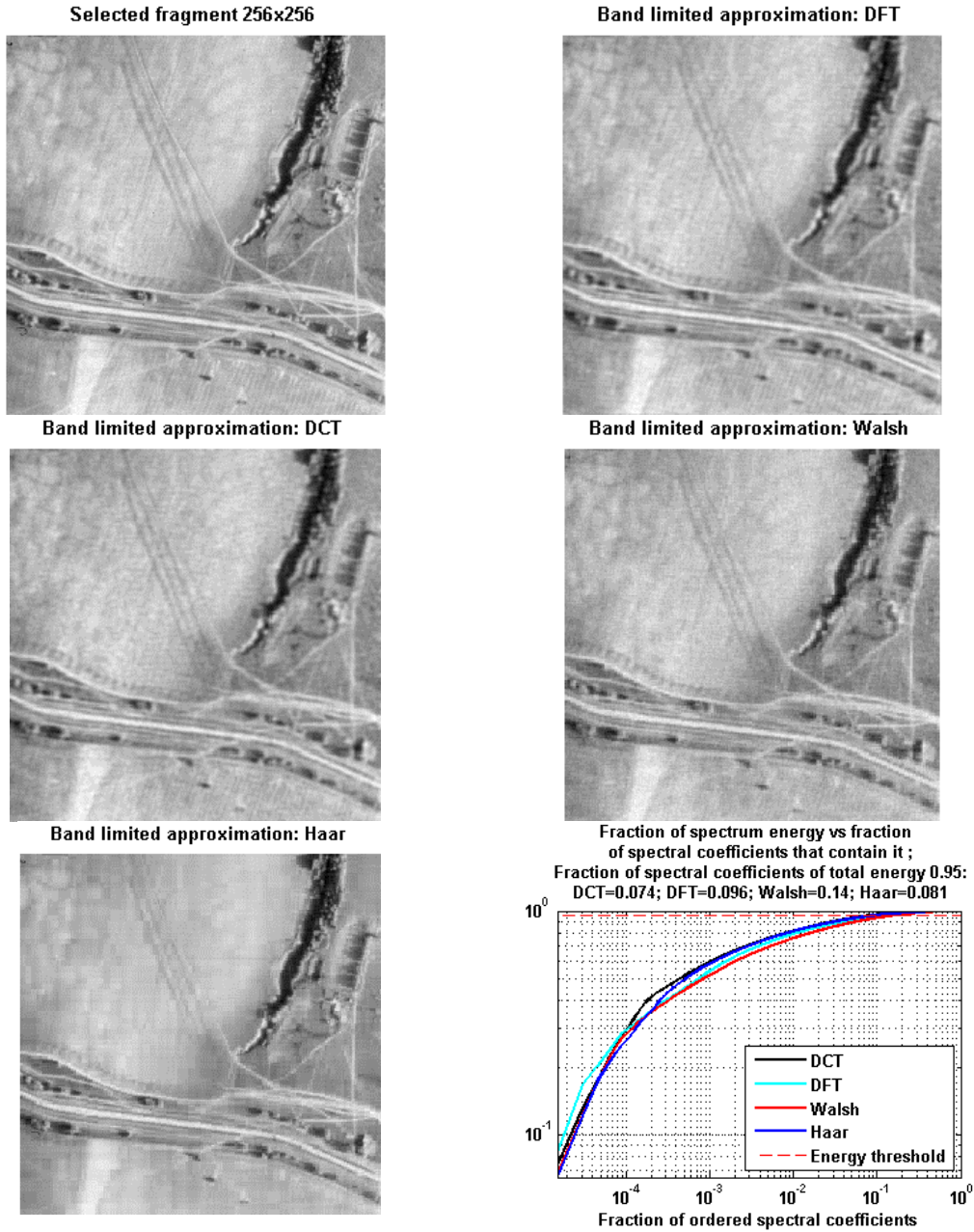

Figure 1. Illustration of the energy compaction capability of Discrete Fourier, Discrete Cosine, Walsh and Haar transforms for a test image shown in the upper left corner. The rest of images show band-limited approximations of the test image in the domain of corresponding transforms for the approximation error variance equal to 5% of image variance. Graphs on the plot in the bottom right corner present energy contained in fractions of transform coefficients for different transforms.

It is noteworthy to mention that introducing of DCT in [7] was motivated not by this reasoning but by the finding that basis functions of DCT provide a good approximation to the eigenvectors of the class of **Toeplitz matrices** with entries $\{\mu_{k,l} = \rho^{|k-l|}\}, \rho = 0.9; k,l = 0,1,...$, i.e. DCT can be considered as a good approximation to the Karhunen-Loeve Transform for signals with the Toeplitz correlation matrix. In fact, the above-mentioned fundamental reason for good energy compaction capability of DCT is hidden in the even symmetry of the Toeplitz matrix.

In the family of orthogonal transforms, DFT and DCT occupy a special place. DFT and DCT are two versions of discrete representation of the Integral Fourier Transform, which plays a fundamental role in signal and image processing as a mathematical model of wave propagation, signal transformations and image formation in imaging systems ([ 16], **Error! Reference source not found.**], [ 19], [ 20]). Thanks to this, DFT and DCT are applicable in much wider range of applications than the other fast transforms (see Table 1).

In conclusion of this section, come back to the second common feature of fast transform, the availability of FFT type fast algorithms for all of these transforms. Initially, fast Fourier transform algorithms were developed only for signals with the number of samples equal to an integer power of two. These are the fastest algorithms. At present, this limitation is overcome, and fast Fourier transform algorithms exist for arbitrary numbers of signal samples. 2D and 3D transforms for image and video applications are built as separable, i.e. they work separably in each dimension. Note that the transform separability is the core idea for fast transform algorithms. All FFT type transform algorithms are based on representation of signal sample and transform coefficient indices as multidimensional numbers, i.e. numbers represented by multiple digits (sub-indices), and on splitting the transform into a separable sequence of shorter 1D transforms over each of the sub-indices.

The data on computational complexity of the transform fast algorithms are collected in Table 2 placed in the Appendix. These data are more or less commonly known. What is not as widely known, it is the existence of the so called ***"pruned" algorithms*** for the cases, when transform input data contain substantial fraction of zero samples and/or one does not need to compute all transform coefficients ([ 14] - [ 28]). These algorithms are useful, in particular, in numerical reconstruction of holograms and in interpolation of sampled data by means of ***transform zero padding*** ([ 19], [ 29], [ 30]).

In some applications it is advisable to apply transforms locally in a window, which slides over the image pixel by pixel, rather than globally to entire image frames. For such application, Discrete Cosine and Discrete Fourier Transforms have an additional very useful feature. Computing these transforms in sliding windows can be carried out recursively: signal transform coefficients at each window position can be found by quite simple modification of the coefficients found at the previous window position. Per pixel computational complexity of such recursive computation is proportional to the window size rather than to the product of the window size and its logarithm, which is the computational complexity of the fast transforms ([ 19], [ 30], [ 31],[ 32]).

The rest of the paper is arranged as following. In Sect. 2, applications of fast transforms for image data compression are reviewed. In Sect. 3 we proceed to transform methods for image restoration and enhancement. In Sect. 4 we show how DFT and DCT can be applied for image perfect, i.e. error less, resampling. And finally in Sect. 5 we address the usage of fast transforms for solving a specific task of image resampling, the task of image recovery from sparse and irregularly taken samples, and, in particular, the compressive sensing approach to this problem.

## 2. Image compression

### *2.1. Dilemma: compressive discretization or data compression*

In this section we address the issue of image digitization, i.e. converting continuous signals from image sensors into digital signals that can be stored and processed in digital computers. Obviously, image digitization should be carried out in such a way as to provide as compact image digital representation as possible for a given quality of image reproduction. In general, this can be done similarly to what we do when we describe the world verbally. In order to do this one should create a look-up table of all possible "representative" images that one wants and can discriminate one from another and perform image digitization by means of assigning to each particular image an index in the look-up-table of a representative image, which is identical to it for the user. This index is the image digital descriptor. Image reconstruction from its digital descriptor will be then done by replacement of the image by its representative image taken from the look-up-table according to the index of the image.

However, such general procedure is feasible only in exceptional cases, when the volume of the image look-up-table can be kept reasonably small. In waist majority of real situations, the number of possible different images is immensely high. To give an estimate of its order of magnitude, consider the number of images of a medium quality, say of 500x500 pixels with 24 bits per pixel. This number is $256^{(3*250000)}$, which apparently exceeds the number of elementary particles in the Universe. For this reason, a two-step image digitization is almost ubiquitous in the imaging engineering: (i) discretization, i.e. converting image sensor signal into a set of real numbers, followed by (ii) scalar quantization, i.e. rounding off, of those numbers to a set of fixed quantized values ([ 14], [ 20]).

Image discretization can, in general, be treated mathematically as obtaining coefficients of image signal expansion over a set of ***discretization basis functions***. In order to make the set of these representation coefficients as compact as possible, one should choose discretization basis functions that secure the least number of the signal expansion coefficients sufficient for image reconstruction with a required quality. We will call this general method of signal discretization "***Compressive discretization***" because it secures the most compact discrete representation of signals, which cannot be further compressed. Note that this term should not be confused with terms "***Compressive sensing***" and "***Compressive sampling***", that gained popularity in recent years ( [ 33]-[ 40]). We will discuss the "compressive sensing" approach later in Sect. 5.

Compressive discretization is a gold standard for discretization. However, in reality in vast majority of imaging devices image discretization is implemented as image spatial sampling at nodes of a uniform rectangular sampling lattice using ***sampling basis functions***, which are formed from one ***"mother" function*** by its shifts by multiple of fixed intervals called ***"sampling" intervals***. In all these cases, compressive discretization is achieved as a two-step procedure via sampling and subsequent compression of sampled data. There are only a few non-sampling discretization methods such that ***Coded Aperture Imaging*** and *Magnetic Resonance Imaging* (MRI). In coded aperture imaging, images are sensed through binary masks that implement binary basis functions. In MRI, measured are coefficients of Fourier series expansion of sensor data.

The theoretical foundation of image sampling is the **sampling theorem**. The traditional formulation of the sampling theorem states that signals with Fourier spectrum limited with bandwidth **B** can be perfectly restored from their samples taken with sampling interval $\Delta x = 1/\mathbf{B}$, commonly called the *Nyquist sampling interval* ([ 41], [ 42], [ 43]).

In reality no continuous signal is band-limited, and the image sampling interval is defined not through specifying, in one or another way, of the image bandwidth, but directly by a requirement to sufficiently good reproduction of smallest objects and borders of larger objects present in images. The selected in this way image sampling interval $\Delta x$ specifies the image *base band* $\mathbf{B} = 1/\Delta x$.

Since small objects and object borders usually occupy relatively small fraction of the image area, vast portions of images are oversampled, i.e. sampled with redundantly small sampling intervals, which results in excessively large total number of image samples, or pixels. Hence, substantial compression of image sampled representation is possible. This is usually implemented in digital processors by means of applying to the image sampled representation a discrete analog of the general compressive discretization, i.e. through image expansion over a properly chosen set of discrete optimal bases functions and limitation of the amount of the expansion coefficients. This is exactly what is done in all transform methods of image compression.

## *2.2. Transform methods of image compression*

As it was already mentioned in the introduction, needs of image compression were the primary motivations of digital image processing at its early years. Being started in 1950-th from various predictive coding methods (a set of good representative publications can be found in [ 44]), by the beginning of 1970-th image compression research began concentrating mostly on what is called "*transform coding*" ([ 45], [ 46]), which literally implements the compressed discretization principle.

The principles of image transform coding and reconstruction is illustrated by the flow diagrams sketched in

Figure 2. According to these diagrams, set of image pixels is first subjected to a fast orthogonal transform. Than low intensity transform coefficients are discarded, which substantially reduces the volume of data. This is the main source of image compression. Note that discarding certain image transform coefficients means replacement of images by their band-limited, in terms of the selected transform, approximations. The remaining coefficients are subjected, one by one, to optimal non-uniform scalar quantization that minimizes the average number of quantization levels of transform coefficients. Finally, quantized transform coefficients are entropy encoded to minimize the average number of bits per coefficient. For image reconstruction from the compressed bit stream, the stream must be subjected to corresponding inverse transformations.

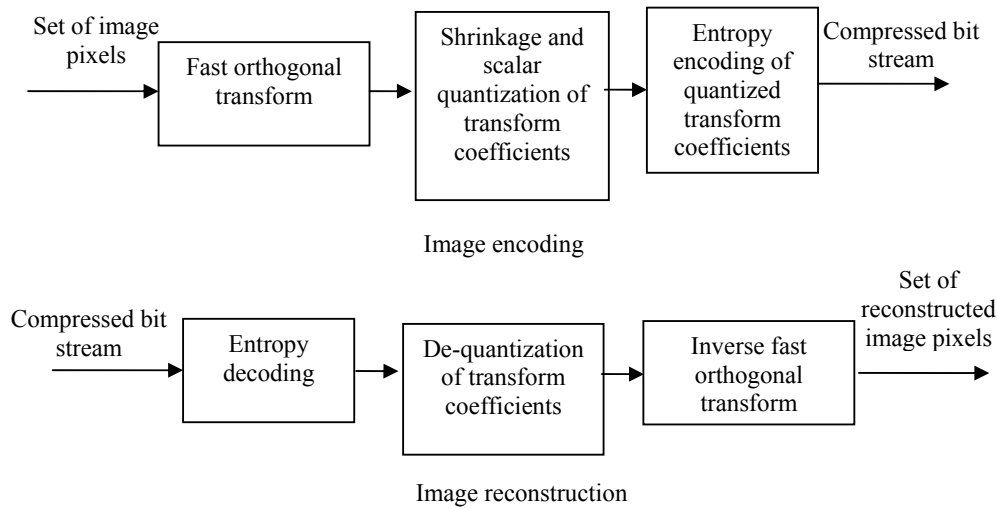

Figure 2. Flow diagrams of image transform coding and reconstruction

In 1970-th, activities of researches were aimed mostly at invention, in addition to known at the time Discrete Fourier, Walsh-Hadamard and Haar Transforms , new transforms that have fast computational algorithms and improved energy compaction capability ([ 3], [ 4], [ 5]).

This transform invention activity gradually faded after introduction, in a short note, of the Discrete Cosine transform ([ 7]), which, as it was already mentioned, was finally recognized as the most appropriate transform for image compression due to its superior energy compaction capability and the availability of the fast algorithm. But the true breakthrough happened, when it was realized that much higher image compression efficiency can be achieved, if transform coding is applied not to entire image frames but block wise ([ 45], [ 46]).

Images usually contain many objects, and their global spectra are a mixture of object spectra, whereas spectra of individual image fragments or blocks are much more specific and this enables much easier separation of "important" (most intense) from "unimportant" (least intense) spectral components. This is vividly illustrated in Figure 3, where global and block wise DCT power spectra of a test image are presented for comparison.

The titles of image spectra on this figure provide numerical data on sparsity on the energy level 0.95 of global and, on average, of block spectra, i.e. on fraction of spectral coefficients that contain 95% of the total spectrum energy. Although, as one can see on the figure, sparsity of DCT global spectrum and, on average, of local spectra are of the same order of magnitude (7.4% and 7.1%, correspondingly), spectra of individual fragments are certainly more specific and their components responsible for small objects and object borders have much higher energy than in the global spectrum and will not be lost when 95% of the most intense components are selected.

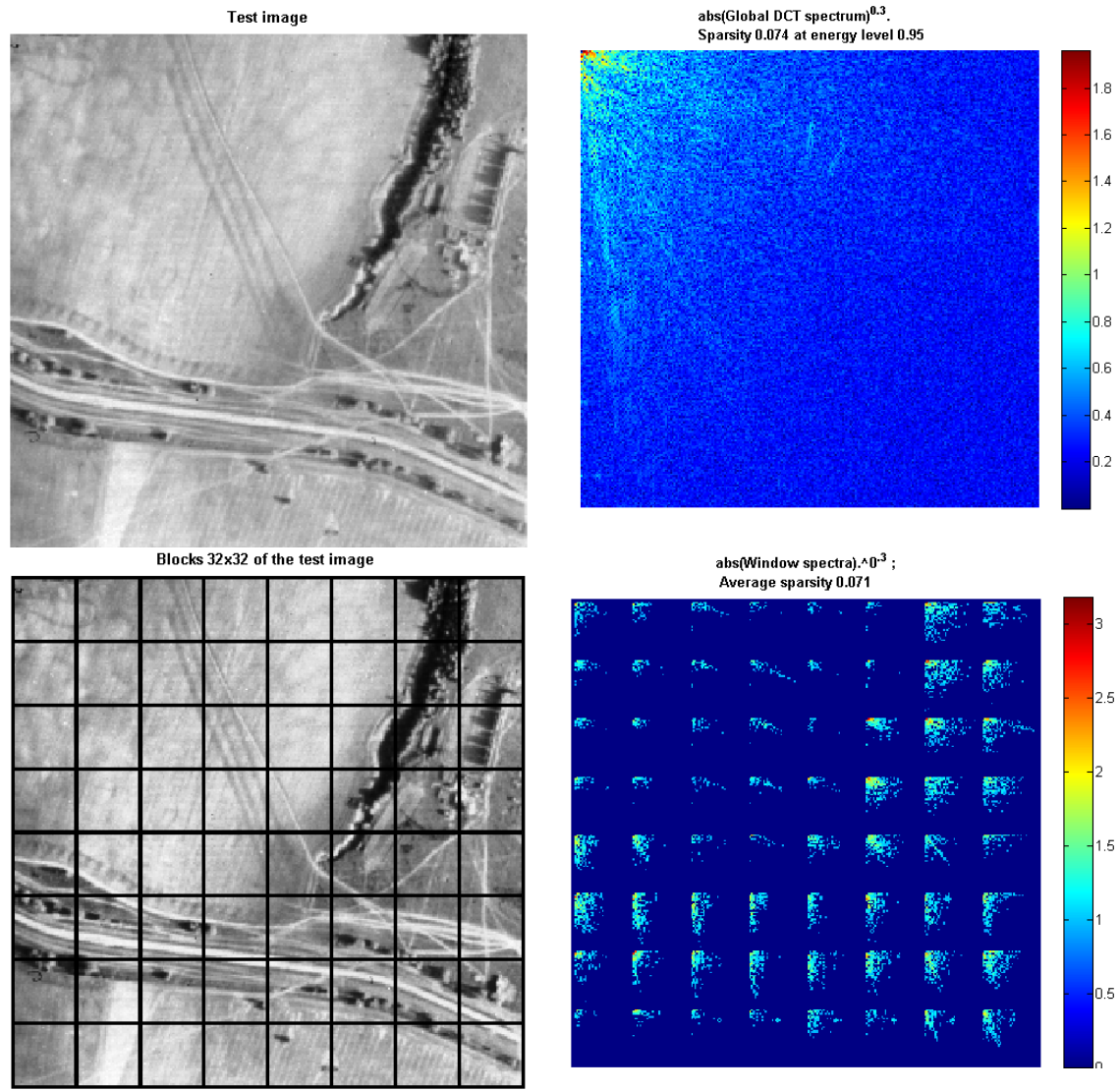

Figure 3. Test image (upper left), its 32x32 pixel blocks (bottom left) and corresponding global (upper right) and local (block-wise, bottom right) spectra DCT shown color coded according to color bars.

Block wise image DCT transform compression proved to be so successful, that, by 1992, it was put in the base of the image and video coding standards "JPEG" and "MPEG" ([ 47] - [ 50]), which eventually resulted in the revolution in the industry of photographic and video cameras and led to the emergence of digital television as well ([ 51]).

Though the standard does not fix the size of blocks, blocks of 8x8 pixels are used in most cases. Obviously, the block transform compression may, in principle, be more efficient, if the block size is adaptively selected in different areas of images, hence the use of variable size of blocks was discussed in a number of publications (see, for instance, [ 52]).

The main factor that limits image compression capability of DCT transform block coding are "***blocking effects***", artificial discontinuities that may appear at the borders of blocks in the result of truncation and quantization of their transform coefficients. The readers may have seen these characteristic effects watching digital cable or internet TV when interferences happen in the communication channel.

The desire to avoid these artifacts motivated the advancement of "Lapped" transforms ([ 71]), which are applied block-wise with the half block size overlap. Even better alternative is wavelet transform coding. It does not require dividing images into blocks at all and achieves very good compression capability thanks to the combination of "local" and "global" sensitivities of wavelet transform coefficients. Multi resolution capability of wavelets used as coding transforms enables, in addition, flexible change of image resolution depending on the image transmission channel capacity. Wavelet transform coding was put in the base of JPEG2000 image compression standard. Although JPEG 2000's improvement in compression performance relative to the original JPEG standard is actually rather modest, it can very efficiently handle a very large range of effective bit rates. For example, to reduce the number of bits for a picture below a certain amount, the advisable thing to do with the first JPEG standard is to reduce the resolution of the input image before encoding it. That is unnecessary when using JPEG 2000, because JPEG 2000 already does this automatically through its multi-resolution decomposition ([ 51]).

## 3. Transform domain filters for image restoration and enhancement

### *3.1. Transform domain scalar Empirical Wiener Filters*

With emergence of image transform compression methods it was realized that transforms represent a very useful tool for image restoration from distortions of image signals in imaging system and for image enhancement as well ([ 53], [ 54]). The principle is very simple: for image perfecting, image transform coefficients are modified in a certain way and then the image is reconstructed by the inverse transform of the modified coefficients (Figure 4). This way of processing is called ***transform domain filtering***.

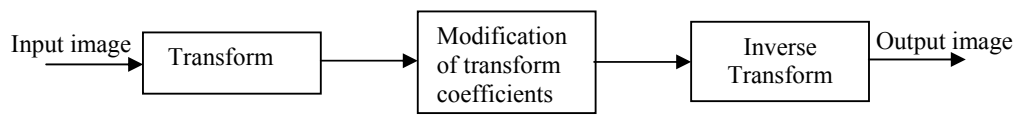

Figure 4. Flow diagram of transform domain filtering

Two options for the modification of image transform coefficients are usually considered:

- Modification of absolute values of transform coefficients by a non-linear transfer function. Usually it is a function that compresses the dynamic range of transform coefficients. Such modification redistributes the coefficients' intensities in favor of less intensive coefficients and results in contrast enhancement of image small details and edges. One of the most simple and practical is ***P-law nonlinear modification*** by a function ***y=x^P***, with 0≤***P***≤1.

- Multiplication of transform coefficients by scalar weight coefficients. This way of processing is called transform domain "*scalar filtering*" ([ 53]).

For defining optimal scalar filter coefficients, a Wiener-Kolmogorov ( [ 55],[ 56]) approach of minimization of mean squared filtering error is used and thus filters implement the *empirical Wiener filtering* principle ([ 19], [ 20], [ 30]).

Three modifications of the filters based on this principle are: (i) Proper *Empirical Wiener Filters*, (ii) *Signal Spectrum Preservation Filters* and (iii) *Rejective Filters* ([ 20]). For image denoising from additive signal independent noise and image deblurring, they are defined in Fourier or DCT domains and modify input image spectra $Sp_{inp}(r,s)$ at each spectral component $(r,s)$ (in denotations of Table 2, Appendix) according to the equations, correspondingly:

*Empirical Wiener Filter*:

$$Sp_{out}(r,s) = \frac{1}{ISFR(r,s)} \max\left(0, \frac{|Sp_{inp}(r,s)|^2 - |Sp_{noise}(r,s)|^2}{|Sp_{inp}(r,s)|^2}\right) Sp_{inp}(r,s) \; ; \qquad (1)$$

*Signal Spectrum Preservation Filter*:

$$Sp_{out}(r,s) = \frac{1}{ISFR(r,s)} \max\left(0, \frac{|Sp_{inp}(r,s)|^2 - |Sp_{noise}(r,s)|^2}{|Sp_{inp}(r,s)|^2}\right)^{1/2} Sp_{inp}(r,s) \; ; \qquad (2)$$

*Rejective Filter*:

$$Sp_{out}(r,s) = \begin{cases} \frac{Sp_{inp}(r,s)}{ISFR(r,s)}, & \text{if } |Sp_{inp}(r,s)|^2 > |Sp_{noise}(r,s)|^2 \\ 0, & \text{otherwise} \end{cases}, \qquad (3)$$

where $Sp_{out}(r,s)$ is spectrum of output image, $|Sp_{noise}(r,s)|^2$ is power spectrum of additive noise, assumed to be known or to be empirically evaluated from the input noisy image ([ 19], [ 20]), $ISFR(r,s)$ is the Imaging System Frequency Response assumed to be known, for instance, from the imaging system certificate.

As one can see in these equations, all these filters eliminate image spectral components that are less intensive than those of noise and the remaining components are corrected by the *"inverse" filter* with frequency response $1/ISFR(r,s)$. Division of image spectra by $ISFR(r,s)$ compensates image blur in the imaging system due to imperfect $ISFR(r,s)$, i.e. it performs image deblurring.

In addition, empirical Wiener filter modifies image spectrum through de-amplification of image spectral components according to the level of noise in them; spectrum preservation filter modifies amplitude of the image spectrum as well by making it equal to a square root of image power spectrum empirical estimate as a difference between power spectrum of noisy image and that of noise. Rejective filter does not modify remaining, i.e. not rejected, spectral components at all. In some publications, image denoising using the spectrum preservation filter is called "*soft thresholding*" and the rejective filtering is called "*hard thresholding*"([ 57], [ 58]).

Versions of these filters are filters that combine image denoising-deblurring and image enhancement by means of *P*-law nonlinear modifications of the corrected signal spectral components

$$Sp_{out}(r,s) = \begin{cases} \dfrac{1}{ISFR(r,s)|Sp_{inp}(r,s)|^{(1-P)}} Sp_{inp}(r,s), & \text{if } |Sp_{inp}(r,s)|^2 > |Sp_{noise}(r,s)|^2 \\ 0, & \text{otherwise} \end{cases} \qquad (4)$$

where *P* is a parameter ($0 \leq P \leq 1$) that controls the degree of spectrum dynamic range compression and, by this, the degree of enhancement of low intensity image spectral components ([ 20]).

If solely image denoising from additive noise is required, the same filters can be implemented in other transform domains (i.e. Walsh, Haar, or whatever); one needs to only remove from their definitions (Eqs. 1-4) the imaging system frequency response $ISFR(r,s)$, which in meaningful only for DFT and DCT as discrete representation of integral Fourier transform. For instance, in Sect. 3.2 their versions for image denoising filtering in domain of wavelet transform is discussed.

The described filters implemented in DFT and DCT and even in Walsh transform domains proved to be very efficient in denoising of so called **narrow band noises**, i.e. noises, such as "*moirè noise*" or "*banding noise*", that contain only few nonzero components in its spectrum in the selected transform ([ 14], [ 19], [ 20]). Some illustrative examples are given in Figures 5 and 6.

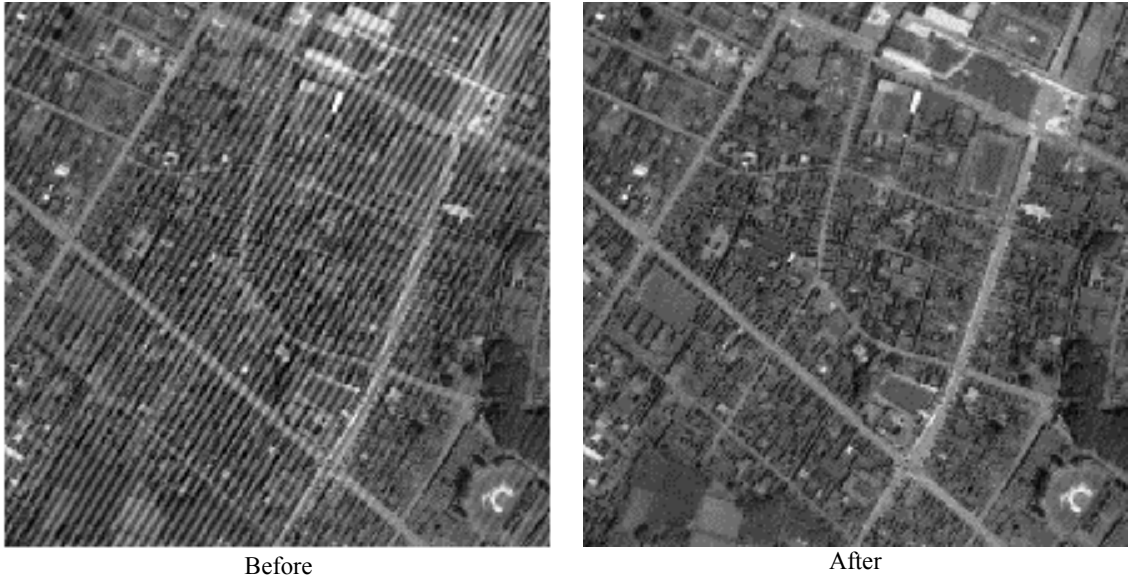

Before    After

Figure 5. Image cleaning from moiré noise

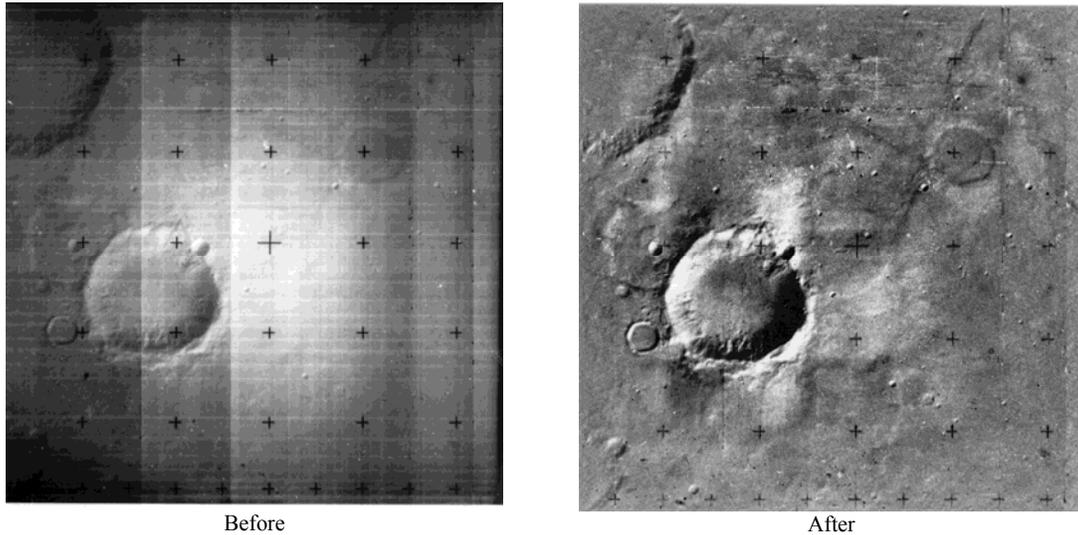
Before                                       After

Figure 6. Cleaning a Mars satellite image from banding noise from by means of separable (row-wise and column-wise) high pass rejective filtering ([ 59],[ 60]).

However, for image cleaning from wide band, or "white", noise, above transform domain filtering applied globally to entire image frames is not efficient. In fact it can even worsen the image visual quality. As one can see from an illustrative example shown in Figure 7, filtered image loses its sharpness and contains residual correlated noise, which is visually more annoying than the initial white noise.

The reason for the inefficiency of global transform domain filtering is the same as for the inefficiency of the global transform compression, which was discussed in Sect. 2.2: important low intensity image spectral components are hidden in global image spectra behind wide band noise spectral components. And the solution of this problem is the same: replace global filtering by local fragment wise filtering. This idea is implemented in *local adaptive linear filters*. These filters, being applied locally, are becoming local adaptive because their frequency responses are determined, according to Eqs. 1-4, by local spectra of image fragments within the filter window. Local adaptive linear filters date back to mid-1980-th and were refined in subsequent years ([ 15], [ 20], [ 30], [ 61]-[ 67]). As a theoretical base for the design and optimization of local adaptive filters, local criteria of processing quality were suggested ([ 68], [ 69]).

It is instructive to note that the evolution of human vision came up to a similar solution. It is well known that, when viewing image, human eye's optical axis permanently hops chaotically over the field of view and that the human visual acuity is very non-uniform over the field of view. The field of view of a man is about 30°. Resolving power of man's vision is about 1′. However such a relatively high resolving power is concentrated only within a small fraction of the field of view that has size of about 2° (see, for instance, [ 70]), i.e. the area of the acute vision is about 1/15-th of the field of view. For images of 512x512 pixels this means window of roughly 35x35 pixels.

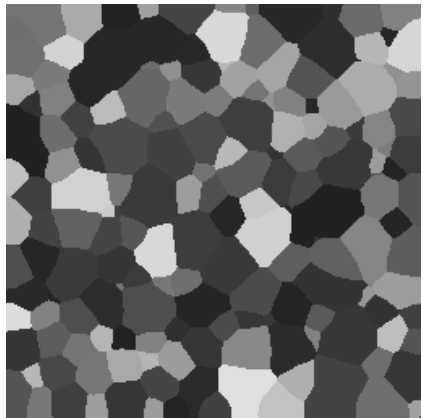
Test image

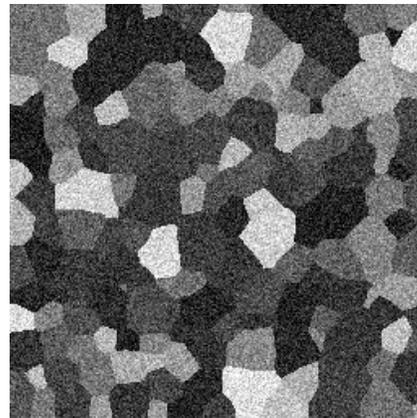
Noisy test image, Noise standard deviation is 25, in units of image range 0-255.

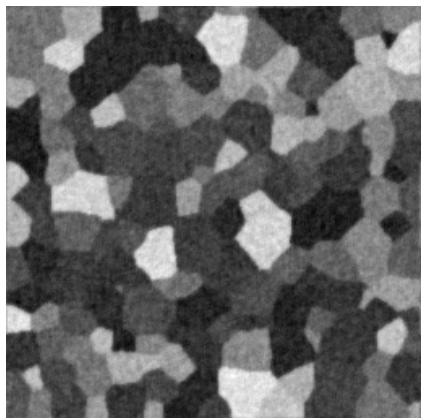
Filtered image

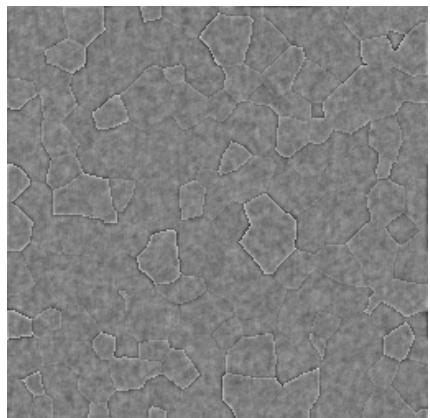
Difference between original test and filtered images. Standard deviation 10.8

Figure 7. Denoising of a piece-wise constant test image using empirical Wiener filter applied to the entire image

The most straightforward way to implement local filtering is to do it in hopping non-overlapping windows, just as human vision does. This is exactly the way of processing implemented in the transform image coding methods. However "hopping window" processing, being very attractive from the computational complexity point of view, suffers from above mentioned "*blocking effects*", artificial discontinuities that may appear, in the result of processing, at the borders of the hopping window. Obviously, the ultimate solution of the "blocking effects" problem would be processing in a sliding window that scans image pixel by pixel in a regular way row-wise/column/wise.

Thus, local adaptive linear filters work in a transform domain in a sliding window and, at each window position, modify, according to the type of the filter defined by Eqs. 1-4, transform coefficients of the image fragment within the window, and then compute an estimate of the window central pixel by means of the inverse transform of the modified transform coefficients [ 67]. As an option, accumulation of estimates of the all window pixels overlapping in the process of image scanning by the filter window is possible as well ([ 72], [ 73], [ 74]). As for the transform for local adaptive filtering, DCT proved to be the best choice in most cases. Figures 8 gives an example of local adaptive filtering for image deblurring, denoising and enhancement.

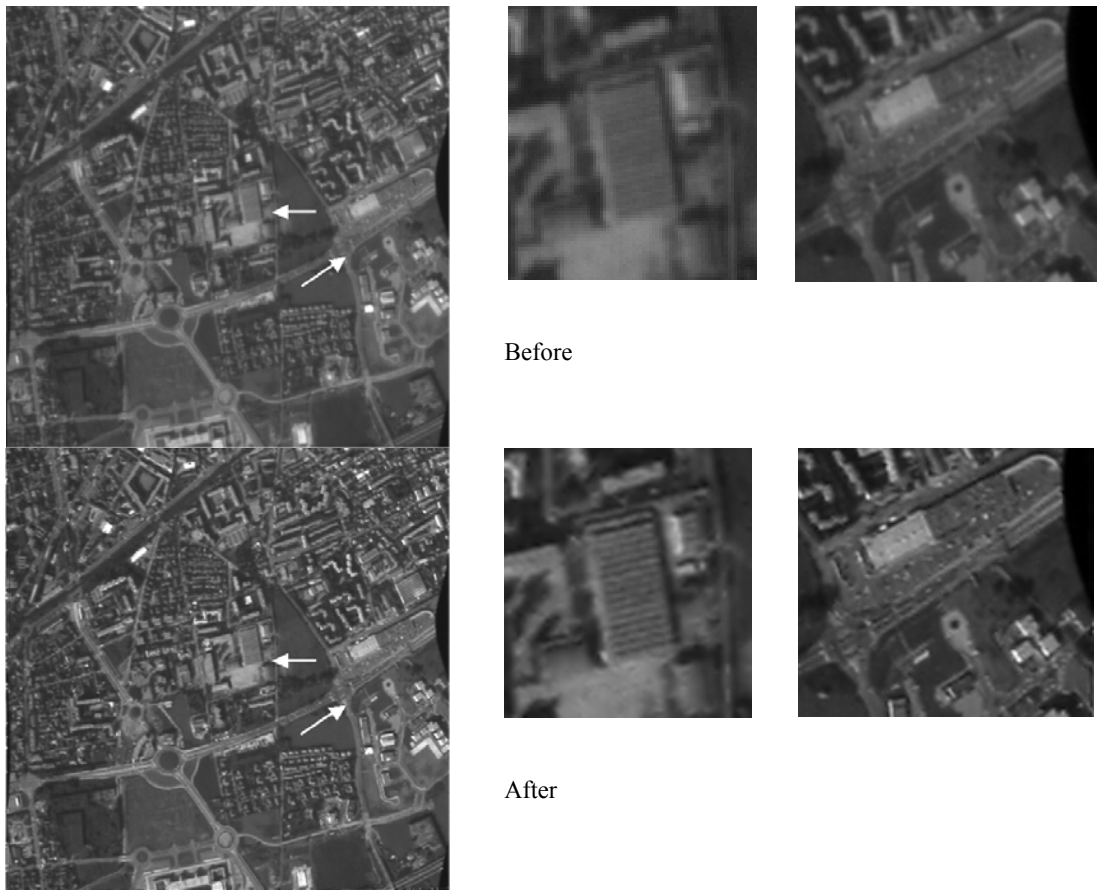

Before

After

Figure 8. Denoising and deblurring a satellite image by means of local adaptive filtering. Top row: raw image and its magnified, for better viewing, fragments marked by arrows; bottom row: resulting image and its corresponding magnified fragments

Local adaptive transform domain filters can be used for perfecting color and multicomponent images and videos as well. In this case, filtering is carried out in the corresponding multi-dimensional transform domains, for instance, domains of 3D (two spatial and one color component coordinates) DCT spectra of color images or 3D DCT spectra of a sequence of video frames (two spatial and one time coordinates). In the latter case filter 3D window scans video frame sequence in both spatial and time coordinates. As one can see from Figures 9 and 10, the availability of the additional third dimension substantially adds to the filter efficiency ([ 75], [ 76]).

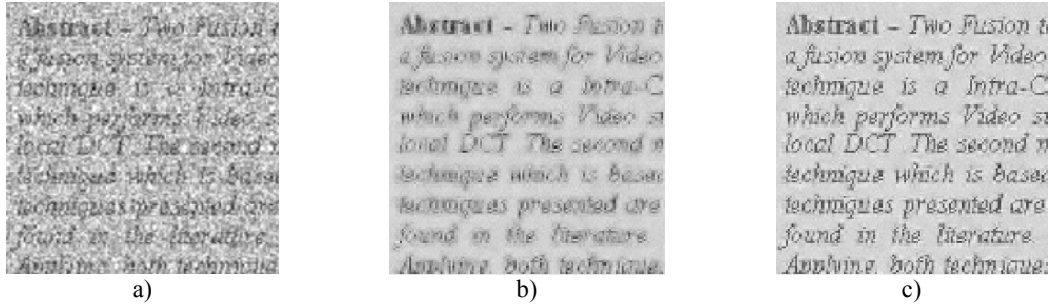

a)                                        b)                                        c)

Figure 9. 2D and 3D local adaptive filtering of a simulated video sequence: a) one of noisy frames of a test video sequence (image size 256x256 pixels); b) a result of 2D frame wise local adaptive DCT domain empirical Wiener filtering (filter window 5x5 pixels); c) a result of 3D spatial and temporal local adaptive DCT domain empirical Wiener filtering (filter window 5x5 pixels x5 frames)

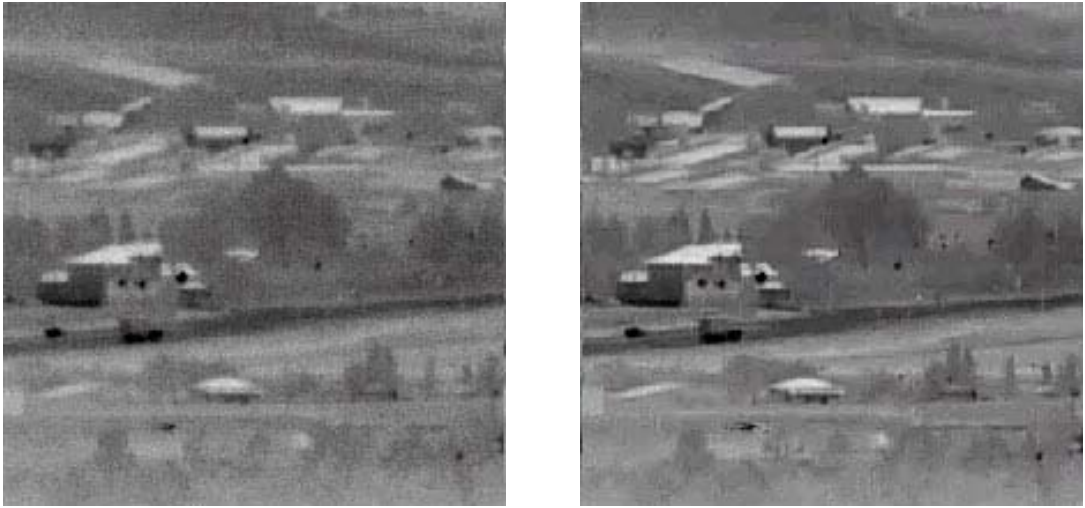

Figure 10. 3D local adaptive DCT domain empirical Wiener filtering for denoising and deblurring of a real thermal video sequence: a frame of the initial video sequence (left) and a frame of a restored video sequence (right). Filter window is 5x5 pixelsx5 frames; image size is 512x512 pixels.

Certain further improvement of image denoising capability can be achieved if, for each image fragment in a given position of the sliding window, similar, according to some similarity measure, image fragments over the whole image frame are found and included in the stack of fragments, to which the 3D transform domain filtering is applied (2D space coordinates and index of fragments in the stack). This approach in its simplest form was coined the name "***non-local means***" filtering ([ 77], [ 78], [ 79]). Such a "non-local" 3D DCT domain filtering is described in Ref. [ 86]) and is recognized in Ref. [ 38] as being among the best performing algorithms for image denoising. Similar method was known much earlier as the *correlational averaging* and was used for high resolution electrocardiography and processing of microscopic images and electron micrographs (see, for instance, [ 80]-[ 85]).  It assumes finding similar image fragments by their cross-correlation coefficient with a fragment chosen as a template and averaging of found in  this way similar fragments (in terms of 3D filtering, this means leaving only the dc component of the 3D spectra in the third dimension).

In assessment of this "non-local" denoising method one should, however, take into account that the correlational averaging is prone to produce artifacts due to false image fragments that may be erroneously taken to the stack just because of the presence of noise that has to be cleaned ([ 87]).

The image restoration efficiency of the *local transform domain filtering* depends on the appropriate selection of the filter window size. Obviously, it will be higher, if the window size is optimized adaptively at each window position. To this goal, filtering can be, for instance, carried out in windows of multiple sizes and, at each window position, the best filtering result should be taken as a signal estimate at this position using methods of statistical tests such as, for instance, "*intersection of confidence intervals method*" described in Ref.[ 88]. Another option in multiple window processing is fusing filtering results obtained for different windows. This can be done by their weighted summation using linear methods of data fusion (see for instance [ 89]), or by taking their median or another robust statistical estimate used in non-linear filters ([ 19]). All this, however, increases correspondingly the computational complexity of the filtering.

In conclusion of this section note that DFT and DCT spectra of image fragments in sliding window processing form the so called "*time - frequency representation*" of signals, which can be traced back to 1930-40-th to D. Gabor and E. Wigner and works on "*visible speech*" ([ 90], [ 91], [ 92]). For 1D-signals, the dimensionality of the "time-frequency representation" is two, for images it is three. Such representation is very redundant and this opens an interesting option of applying to it image (for 1D signals) and video (for 2D signals) processing methods for signal denoising and, in general, for signal separation ([ 93]).

### 3.2. *Wavelet shrinkage filters for image denoising.*

In 1990-th, a specific family of transform domain denoising filters, the so called *wavelet shrinkage* filters, gained popularity after publications [ 57], [ 58] ], [ 94], [ 95]). The filters work in the domain of one of wavelet transforms and implement, for image denoising, the above mentioned Signal Spectrum Preservation Filter (Eq. 2) and Rejective Filter (Eq. 3), (except that they do not include the "inverse filtering" component ($1/ISFR$) that corrects distortions of image spectra in imaging systems). As it was already mentioned, the filters were coined in these publications names "*soft thresholding*" and "*hard thresholding*" filters, correspondingly.

As it was already mentioned, basis functions of wavelet transforms (wavelets) are formed by means of a combination of  shifting and scaling of a *"mother" function* and, thanks to this, wavelets combine local representativeness of *shift basis functions* and global representativeness of *"scaled" basis functions* and feature multi-resolution. Thus the wavelet shrinkage filters, being applied to entire image frames, possess both global adaptivity and local adaptivity in different scales, and do not require for the latter specification of the window size, which is necessary for the sliding window filters.

The wavelet shrinkage filters proved to have a good denoising capability. However, a comprehensive comparison of their denoising performance with that of sliding window DCT domain filters reported in Refs. [ 72],[ 73],[ 74] showed that, even when for particular test images the best wavelet transform from a transform set ([ 96]) were selected, moving window DCT domain filters demonstrated in most cases better image denoising results.

The capability of wavelets to represent images in different scales can be exploited for improving the image denoising performance of both families of filters and for overcoming the above mentioned main drawback of sliding window DCT domain filters, the fixed window size that might be not optimal for different image fragments. This can be achieved by means of incorporating sliding window DCT domain (SWTD-) filters into the wavelet filtering structure through replacing soft/hard thresholding of image wavelet decomposition components in different scales by their filtering with SWTD-filters working in the window of the smallest size 3x3 pixels. This idea of hybrid wavelet/SWTD-filters proved to be fruitful in ([ 97], see also [ 67]).

### 3.3. *Local adaptive filtering and wavelet shrinkage filtering as processing of image sub-band decompositions*

Obviously, both sliding window transform domain and wavelet processing are just different implementations of the *scalar linear filtering in transform domain*. There is yet another approach to their unified interpretation that allows gaining a deeper insight into their similarities and dissimilarities. It is based on the notion of signal sub-band decomposition [ 98]. It was shown in Ref. [ 99] (see also [ 30], [ 67]) that both filter families can be treated as special cases of signal *sub-band decomposition* by band-pass filters with point spreads functions defined by the corresponding transform basis functions.

From the point of view of signal sub-band decomposition, the difference between sliding window transform domain and wavelet signal analysis, besides the differences between sub-band filter point spread functions, is arrangement of bands in the signal frequency range. While for sliding window transform domain filtering sub-bands are uniformly arranged within the signal base band, sub-bands of the wavelet filters are arranged in a logarithmic scale. Hybrid wavelet/SWDCT filtering combines these two types of sub-band arrangements: "coarse" sub-band decomposition in a logarithmic scale of wavelet decomposition is complemented with "fine" uniformly arranged sub-sub-bands within each of the wavelet sub-bands in the sliding window DCT filtering of the wavelet sub-bands.

It is curiously to note that this "logarithmic coarse - uniform fine" sub-band arrangement resembles very much the arrangements of tones and semi-tones in music. In Bach's equal tempered scale, octaves are arranged in a logarithmic scale and 12 semitones are equally spaced within octaves ([ 100]). The same idea of "logarithmic coarse - uniform fine" representation is also implemented in representation of numbers in computers by their order and mantissa.

## 4. Image resampling and building "continuous" image models

As it was indicated in the introductory section, Discrete Fourier and Discrete Cosine are two versions of discrete representation of the integral Fourier Transform. Because of this, among applications specific for DFT and DCT are signal and image spectral analysis and analysis of periodicities, fast signal and image convolution and correlation and image resampling and building "continuous" image models ([ 19], [ 20], [ 30]). The latter is associated with the efficiency, with which DFT and DCT can be utilized for fast signal convolution.

Image re-sampling is a key operation in solving many digital image processing tasks. It assumes reconstruction, out of the available sampled image, of an approximation to the original non-sampled image by means of interpolation of available image samples to obtain samples "in-between" the available ones. The most feasible is interpolation by means of digital filtering implemented as digital convolution. A number of convolutional interpolation methods are known, beginning from the simplest and the least accurate nearest neighbor and linear (bilinear, for 2D case) interpolations to more accurate cubic (bicubic, for 2D case) and higher order spline methods ([ 101], [ 102]). When only image rescaling, i.e. down-sampling or up-sampling (subsampling) is required of images available in a JPEG2000 compressed form, the rescaling can be easily performed using, for down-sampling, corresponding lower scales of the encoded images or, for subsampling, zero padding of image wavelet coefficients. This option is however available when the required scaling factor is the same as that of wavelets used for compression (usually, it is a power of 2).

The interpolation accuracy of spline interpolation methods is limited by the spline order: the higher spline order, the higher the interpolation accuracy. The per pixel computational complexity of spline interpolators is proportional to the spline order.

There exists a discrete signal interpolation method that secures error free interpolation of sampled signals specified by finite sets of their samples, i.e. the method that, contrary to other methods, does not add to sampled signals any distortions additional to the primary distortions caused by the signal sampling. This method is the ***discrete sinc-interpolation*** ([ 20], [ 29]). The interpolation kernel of the discrete sinc-interpolation of $N$ samples is the ***discrete sinc-function*** $\text{sincd}\, x = \dfrac{\sin x}{N \sin(x/N)}$. This function is a discrete analog, for sampled signals with a finite number of samples, of the continuous ***sinc-function*** $\text{sinc}\, x = \dfrac{\sin x}{x}$, which, according to the sampling theory, is the ideal interpolation kernel for reconstruction of continuous signals from their samples provided that the number of samples is infinitely large ([ 20]). The interpolation accuracy of the discrete sinc interpolation is limited only by the accuracy of computations.

Note that splines, with growths of their order, tend asymptotically to the discrete sinc-function (assuming cyclic signal extension outside its borders). A similar theorem for continuous splines and continuous sinc-function was proven by A. Aldroubi and M. Unser ([ 101]).

The discrete sinc function has the same number of samples as signals, for interpolation of which it is used. Therefore discrete sinc interpolation through direct convolution in signal domain would require $N$ addition and multiplication operation per output interpolated signal sample, which is obviously prohibitively large. Implementation of the convolution through Fast Fourier transform reduces its computational complexity to the order of $\log N$, which makes discrete sinc interpolation competitive with other less accurate numerical interpolation algorithms.

However, FFT has a substantial drawback in computing convolution: as it implements the cyclic convolution with a period equal to the number $N$ of signal samples, at signal borders it involves in the interpolation signal samples from the signal opposite borders. Therefore it is potentially prone of causing heavy oscillations at signal borders because of a discontinuities, which may happen between signal values at its opposite borders.

This drawback is to a very high degree surmounted, if signal convolution is implemented through DCT ([ 20], [ 30]). As it was already indicated, convolution through DCT is also a cyclic convolution, with a period of $2N$ samples, of signals that are evenly extended to double length by their inversed, in the order of samples, copies. This type of signal extension secures the absence of discontinuities at signal borders and thus eliminates the danger of appearance of heavy oscillations at signal borders that are characteristic for discrete sinc-interpolation through processing in DFT domain.

There are several implementations of DFT/DCT based discrete sinc interpolation. The most straightforward one is DFT/DCT spectrum zero padding. Interpolated signal is generated by applying inverse DFT (or, correspondingly, DCT) transform to its zero padded spectrum. Padding DFT/DCT spectra of signals of $N$ samples with $L(N-1)$ zeroes produces $LN$ samples of the discrete sinc interpolated initial signal with subsampling rate $L$, i.e. with inter-sample distance $1/L$-th of the initial sampling interval. The computational complexity of this method is $O(\log LN)$ per output sample.

The same discrete sinc-interpolated $L$ times subsampled signal can be obtained more efficiently computation-wise by applying to the signal $L$ times the signal ***perfect shifting filter*** ([ 29], [ 103], [ 104], [ 105]) with signal shift, at $k$-th application, by $k/L$ and by subsequent combining the shifted signal copies. The computational complexity of this method is $O(\log N)$ per output sample.

This method is based on the property of **Shifted DFT** (SDFT, Table 2 and [ 106], [ 107], [ 108]): if one computes SDFT of a signal with some value of the shift parameter in signal domain and then computes inverse SDFT with another value of the shift parameter, the result will be a discrete sinc interpolated copy of the signal shifted by the difference between those values of the shift parameter. The DCT based version of this filter ([ 109], [ 110], [ 111], [112]) avoids the danger of objectionable oscillations at signal borders and is recommended for practical use.

Image subsampling using the perfect shifting filter can be employed for creating "continuous" image models for subsequent image arbitrary resampling with any given accuracy ([ 20], [ 113]). It can also be used for computing image correlations and image spectral analysis with sub-pixel accuracy ([ 20], [ 30]). In [ 114] (see also [ 20]), particular examples of using created in this way continuous signal models for converting image spectra from polar coordinate system to Cartesian coordinate system in the direct Fourier method for image reconstruction from projections and for converting (rebinning) of image fan beam projections into parallel projections are described.

The perfect shifting filter is also ideally suited for image sheering in the three pass method for fast image rotation by means of three subsequent (horizontal/vertical/horizontal) image sheerings ([ 115]).

A certain limitation of the above described method for image subsampling using the perfect shifting filter is its capability of subsampling images only with a rate expressed by an integer or a rational number. In some cases this might be inconvenient, as for instance, when the required resampling rate is a value between one and two, say 1.1, 1.2 or alike. For such cases, there exists a third method of signal rescaling with discrete sinc-interpolation. It is based on the general Shifted Scaled (ShSc) DFT (see Table 2), which includes arbitrary analog shift and scale parameters.

Using ShScDFT, one can apply to the image DFT spectrum inverse ShScDFT with the desired shift and scale parameters and obtain a correspondingly shifted and scaled discrete sinc-interpolated image ([ 111], [112]). Similarly, one can arbitrarily shift, scale and rotate image in one step using direct and inverse Shifted Scaled Rotated DFT (see Table 2). This method also implements perfect discrete sinc-interpolation.

Both Shifted Scaled DFT and Shifted Scaled Rotated DFT can be presented as a convolution, and, as such, can be computed using Fast Fourier or Fast Cosine Transforms ([ 20], [ 30]). Thus the computational complexity of the method is $O(\sigma \log N)$ per output signal sample, where $\sigma$ is the scale parameter (resampling rate). As in all other cases, the use of DCT based convolution guaranties the absence of severe boundary effects. Image resizing using Shifted Scaled DFT is especially well suited for numerical reconstruction of digitally recorded color holograms, in which it is required to resize images reconstructed from holograms recorded with different wave length of coherent illumination, when scaling factors may duffer one from another by only a fraction of unit [ 111].

Perfect interpolation capability of the discrete sinc-interpolation through processing in DFT/DCT domains was demonstrated in a comprehensive comparison of different interpolation methods, including spline of order 9 ("Mems531", [ 102], [ 117]), in experiments with multiple 360$^o$ rotations reported in Ref. [ 29] and in experiments with multiple alternative image zooming-in and zooming-out reported in Ref. [112]. The results of experiments illustrated in Figures 11 for image rotation and in Figure 12 for image zooming-in and zooming-out clearly evidence that, in contrast to other methods, discrete sinc interpolation does not introduce any appreciable distortions into interpolated images. In addition note that in the image rotation experiments reported in Ref. [ 29] discrete sinc-interpolation proved to be very competitive with other interpolation methods in terms of computation speed as well.

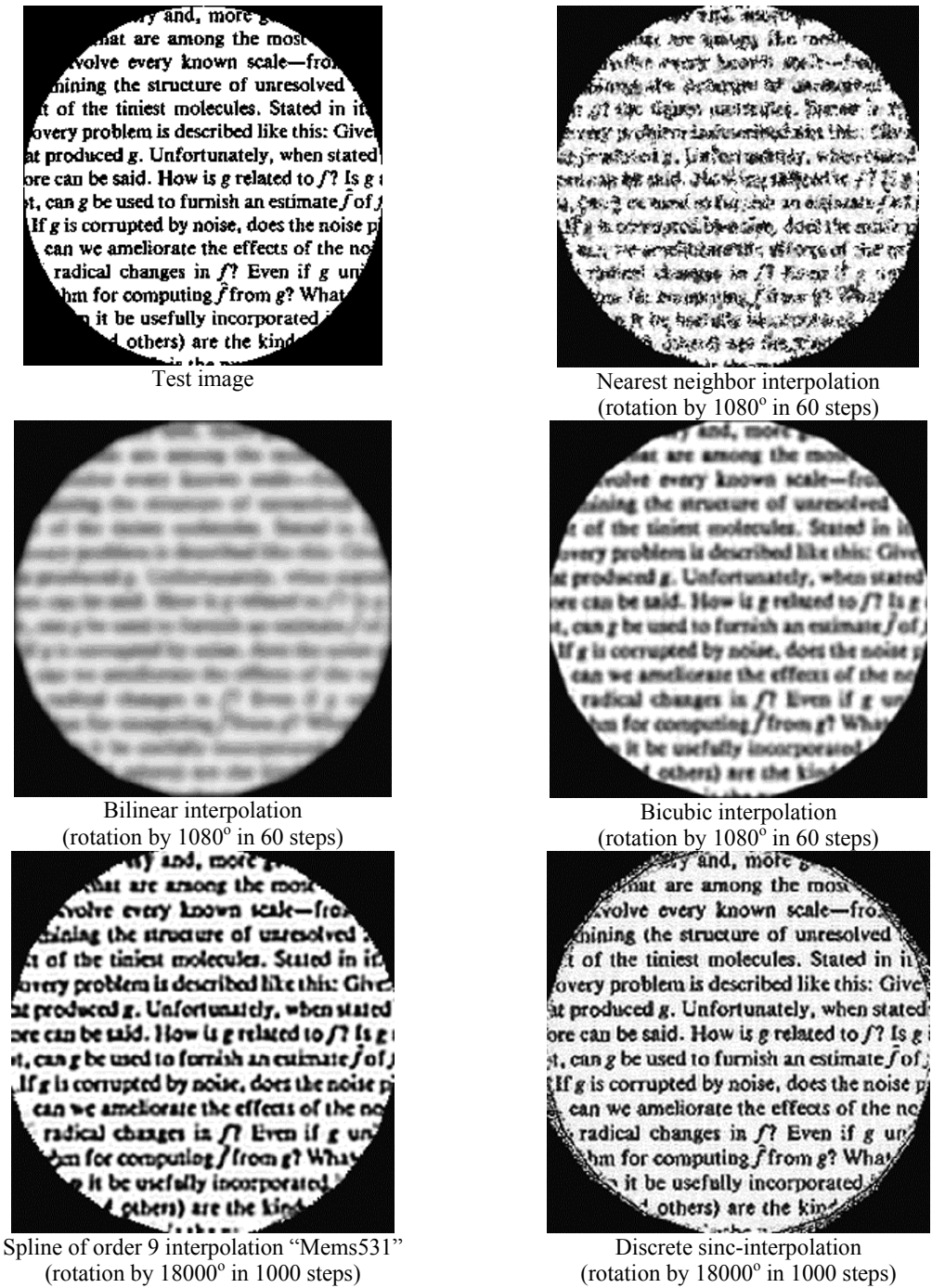

Figure 11. Discrete sinc-interpolation versus other interpolation methods: results of multiple image rotations

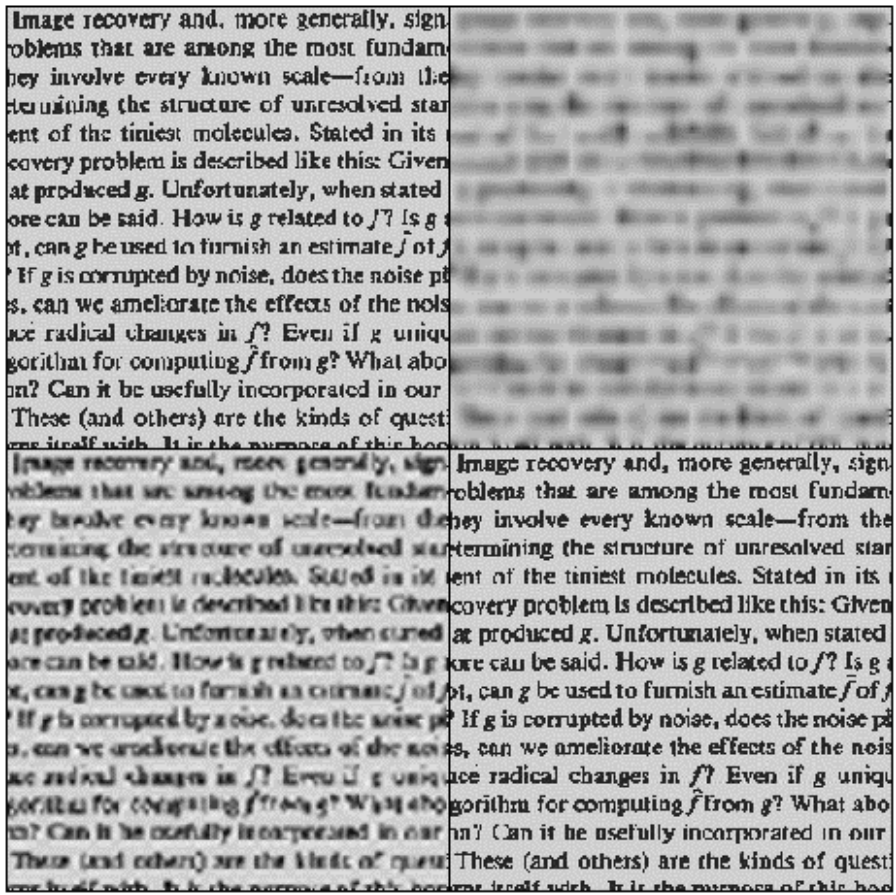

Figure 12. Discrete sinc-interpolation versus bilinear and bicubic interpolations in image 75 iterative zooming-ins/zooming-outs with the scale parameter √2

In some applications, "elastic", i.e. space variant, image resampling is required, when shifts of pixel positions are specified individually for each image pixel. In these cases one can either resample, according to the required pixel displacements, continuous image models generated by above described methods or apply the perfect shifting filter to image fragments in sliding window for evaluating shifted interpolated value of the window central pixel at each window position. Applications of these approaches to imitation of image retrieval through turbulent media (an illustration one can see at [ 120]) and to stabilization and super-resolution of turbulent videos are presented in Refs. [ 121], [ 122] and [ 123].

"Elastic" image resampling in sliding window can be easily combined with above described local adaptive denoising and deblurring ([ 29]). Yet another additional option is adaptive resampling in sliding window by means of switching between the perfect discrete sinc interpolation and another interpolation method better suited for specific image fragments ([ 29], [ 109]). This might be useful in application to very heterogeneous images that contain fragments of both natural images and artificially created images such as binary drawings and printed text. The fact is that the perfect discrete sinc interpolation tends to reproduce the latters with oscillations, which might look visually annoying in displayed images. Using for such image fragments even the simplest nearest neighbor interpolation instead of discrete sinc interpolation will allow to prevent appearance of such objectionable artifacts.

Among the image processing tasks, which involve "continuous" image models, are also signal differentiation and integration, the fundamental tasks of numerical mathematics that date back to such classics of mathematics as Newton and Leibnitz. One can find standard numerical differentiation and integration recipes in numerous reference books, such as, for instance, [ 118] and [ 119]. All of them are based on signal approximation from its samples through Taylor expansion. However, according to the sampling theory, approximation of sampled signal by Taylor expansion is not optimal, and the direct use of these methods for sampled signals may cause substantial errors in cases, when signals contain substantial high frequency components. In order to secure accurate signal differentiation and integration by means of those standard algorithms one must substantially oversample such signals.

The exact solution for the discrete representation of signal differentiation and integration provided by the sampling theory tells that, given the signal sampling interval and signal sampling and reconstruction devices, discrete frequency responses (in DFT domain) of digital filters for perfect differentiation and integrations should be proportional and, correspondingly, inversely proportional to the frequency index ([ 20], [ 30]). This result directly leads to fast differentiation and integration algorithms that work in DFT or DCT domains using corresponding fast transforms with computational complexity $O(\log N)$ operation per signal sample for signals of $N$ samples. As in all cases of digital convolution, realization of the integration and, especially, differentiation filters in DCT domain is preferable because of the above mentioned much lower vulnerability of DCT based convolution to border effects.

The comprehensive comparison of the accuracy of standard numerical differentiation and integration algorithms with perfect DCT-based differentiation and integration is reported in [ 29], [ 124], where it is shown that the standard numerical algorithms perform, in reality, differentiation and integration of not original signals but of their blurred to a certain degree copies. This conclusion is illustrated in Figure 13 on an example of multiple alternative differentiations and integrations, using standard and DCT-based differentiation and integration methods, of a test signal in form of two delta-impulses.

Computational efficiency of the DFT/DCT based interpolation error free discrete sinc-interpolation algorithms is rooted in the use of fast Fourier and Fast DCT transforms. Perhaps, the best concluding remark for this discussion of the discrete sinc interpolation DFT/DCT domain methods would be mentioning that a version of what we call now Fast Fourier Transform algorithm was invented more than 200 years ago by Carl Friedrich Gauss just for the purpose of facilitating numerical interpolation of sampled data of astronomical observation. ([ 125]).

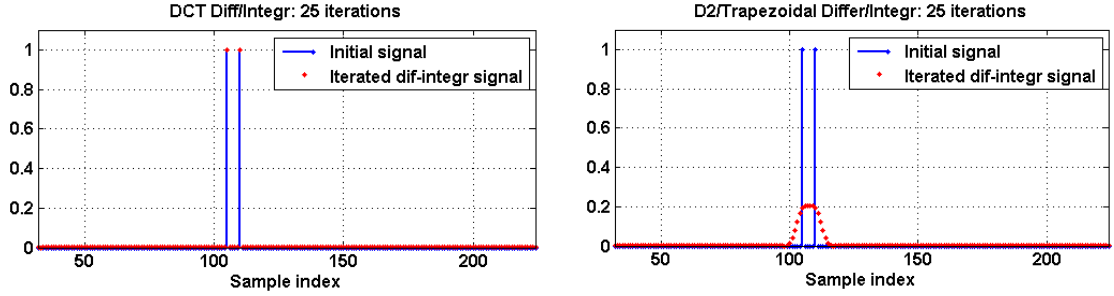

Figure 13. Comparison of results of iterative alternated differentiations and integrations of a test signal (two delta-impulses separated by 4 sampling intervals) using the DCT-based algorithms (left) and using standard numerical algorithms (differentiation filter with point spread function [-0.5, 0, 0.5] and trapezoidal rule integration algorithm). After 25 iterations using DCT based algorithms the tests signal remains unchanged, whereas after the same number of iterations using the standard numerical algorithms two delta impulses became unresolved.

## 5. Image recovery from sparse and irregularly sampled data

### 5.1. Discrete sampling theorem based method

As we mentioned at the beginning of Section 2, image discretization is usually carried out by image sampling at nodes of a uniform rectangular sampling lattice. In this section we address using fast transforms for solving the problem closely associated with the general compressive discretization: how and with what accuracy can one recover an image from its sparse and irregularly taken samples.

There are many applications, where, contrary to the common practice of uniform sampling, sampled data are collected in irregular fashion. Because image display devices as well as computer software for processing sampling data assume using regular uniform rectangular sampling lattice, one needs in all these cases to convert irregularly sampled data to regularly sampled images.

Generally, the corresponding regular sampling grid may contain more samples than it is available, because coordinates of positions of available samples might be known with a "sub-pixel" accuracy, that is with the accuracy (in units of image size) better than $1/K$, where $K$ is the number of available pixels. Therefore one can regard available $K$ samples as being sparsely placed at nodes of a denser sampling lattice with the total amount of nodes $N > K$.

The general framework for recovery of discrete signals from a given set of their arbitrarily taken samples can be formulated as an approximation task in the assumption that continuous signals are represented in computers by their $K < N$ irregularly taken samples and it is believed that if all $N$ samples in a certain regular sampling lattice were known, they would be sufficient for representing those continuous signals ([ 20], [ 126]). The goal of the processing is generating, out of an incomplete set of $K$ samples, a complete set of $N$ signal samples in such a way as to secure the most accurate, in terms of the reconstruction mean square error (MSE), approximation of the discrete signal, which would be obtained if the continuous signal it is intended to represent were densely sampled at all $N$ positions.

Above described discrete sinc-interpolation methods provide band-limited, in terms of signal Fourier spectra, approximation of regularly sampled signals. One can also think of signal band limited approximation in terms of their spectra in other transforms. This approach is based on the ***Discrete Sampling Theorem*** ([ 20], [ 126], [ 127]).

The meaning of the Discrete Sampling Theorem is very simple. Given $K$ samples of a signal, one can, in principle, compute $K$ certain coefficients of a certain signal transform and then reconstruct $N$ samples of a band-limited, in this transform, approximation of the signal by means of inverse transform of those $K$ coefficients supplemented with $(N-K)$ zero coefficients. If the transform has the best, for this particular type of signals energy compaction capability and selected were those non-zero transform coefficients that concentrate the highest, for this particular transform, signal energy, the obtained signal approximation will have the least mean square approximation error. If the signal is known to be band-limited in the selected transform and the computed non-zero coefficients correspond to this band limitation, signal will be reconstructed precisely.

The discrete sampling theorem is applicable to signals of any dimensionality, though the formulation of the signal band-limitedness depends on the signal dimensionality. For 2D images and such transforms as Discrete Fourier, Discrete Cosine, Walsh Transforms, the most simple and quite natural is compact "low-pass" band-limitedness by a rectangle or by a circle sector. For wavelet transforms band-limitedness in terms of transform scale (resolution level) is the most natural.

It was shown in ([ 126], see also [ 20], [ 127]) that for 1D signal and such transforms as DFT and DCT, signal recovery from sparse samples is possible for arbitrary positions of sparse samples. For images the same is true, if band-limitation conditions are separable over the image dimensions. For non-separable band-limitations, such as circle or circle sector this may not be true and certain redundancy in the number of available samples might be required to secure exact recovery of band limited images.

As it was indicated, the choice of the transform must be made on the base of the transform energy compaction capability for each class of images, for which the particular image to be recovered, is believed to belong. The type of the band-limitation must also be based on a priori knowledge regarding the class of images at hand. The number of samples to be recovered is a matter of a priori belief of how many samples of a regular uniform sampling lattice would be sufficient to represent the images for the end user.

Implementation of signal recovery/approximation from sparse non-uniformly sampled data according to the discrete sampling theorem requires matrix inversion, which is, generally, a very computationally demanding procedure. In applications, in which one can be satisfied with image reconstruction with a certain limited accuracy, one can apply for the reconstruction a simple iterative reconstruction algorithm of the Gerchberg-Papoulis ([ 128], [ 129]) type, which, at each iteration step, alternatively applies band limitation in spectral domain and then restores available pixels in their given positions in the image obtained by the inverse transform.

Among reported applications, mention super-resolution from multiple video frames of turbulent video and super-resolution in computed tomography that makes use redundancy of object slice images, in which usually a substantial part of their area is an empty space surrounding the object ([ 126]).

*5.2.     "Compressive sensing": promises and limitations*

The described discrete sampling theorem based methods for image recovery from sparse samples by means of their band-limited approximation in a certain transform domain require explicit formulation of the desired band limitation in the selected transform domain. For 1D signal this formulation requires, for the most frequently used low pass band limitation, specification of only one parameter, the highest index of non-zero signal spectral coefficients. In 2D case formulation of the signal band limitation requires specification of a 2D shape of signal band-limited spectrum. The simplest shapes, rectangle or circular and circle sector ones, may only approximate, with a certain and, possible, considerable redundancy, shapes of real areas occupied by the most intensive image spectral coefficients for particular images. Figure 14 illustrates this on an example of spectral binary mask that indicates, for the test image shown in Figure 3, positions of 7.4% of its DCT spectral coefficients that contain 95% of its spectrum energy. Also shown in Figure 14 is a rectangular outline of this mask, which encompasses 48% of the spectral coefficients.

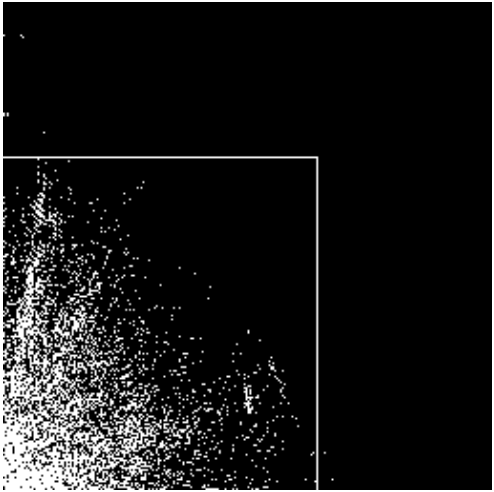

Figure 14. Spectral binary mask (shown white, with the dc component at the left bottom corner) that indicates those 7.4% of components of DCT spectrum of the image in Figure 3 that contain 95% of the total spectrum energy. The rectangular outline of the mask contains 48% of spectral coefficients.

In the cases, when exact character of spectrum band limitation is not known, image recovery from sparse samples can be attempted using the "compressive sensing" approach introduced in Refs. [ 33] - [ 36]. During last recent years, this approach to handling sparse spectral image representation has obtained considerable popularity ([ 37] - [ 40]).

The compressive sensing approach assumes, similarly to the above described image recovery methods, obtaining a band-limited, in a certain selected transform domain, approximation of images as well, but it does not require explicit formulation of the image band-limitation and achieves image recovery from an incomplete set of samples by means of minimization of *L0* norm of the image spectrum in the selected transform (i.e. of the number of signal non-zero transform coefficients), conditioned by preservation in the recovered image its available pixels.

However, there is a price one should pay for the uncertainty regarding the band limitation: the number $M$ of samples required for recovering $N$ signal samples by this approach must be redundant with respect to the given number $K$ of non-zero spectral coefficients: $M = K \log N$ ([ 34]). In real applications, this might be a serious limitation. For instance, for the test image of 256x256 pixels shown in Figure 3 spectrum sparsity (relative number of non-zero spectral components) on the energy level 95% is 0.074, whereas $\log N = \log(256 \times 256) = 16$, which means that the "compressive sensing" approach requires in this case more signal samples ($16 \cdot 7.4\% = 118.4\%$) than it is required to recover. Apparently, for applicability of the compressive sensing approach expected spectra sparsity of images of $N$ pixels should be lower than $1/\log N$.

### 5.3. *Discrete signal band-limitedness and the discrete uncertainty principle*

Signal band-limitedness plays an important role in dealing with both continuous signals and discrete (sampled) signals representing them. It is well known that continuous signals can't be both strictly band-limited and have strictly bounded support. In fact, real continuous signals are neither band-limited nor have strictly bounded support. They can only be more or less densely concentrated in signal and spectral domains. This property is mathematically formulated in the form of the "*uncertainty principle*" ([ 130]):

$$X_{\varepsilon E} \times F_{\varepsilon E} \geq 1 , \quad (5)$$

where $X_{\varepsilon E}$ is interval in signal domain that contains $(1-\varepsilon E)$- fraction of its entire energy, $F_{\varepsilon E}$ is interval in signal Fourier spectral domain that contains $(1-\varepsilon E)$- fraction of signal energy and $\varepsilon E$ is assumed to be sufficiently small.

In distinction to continuous signals, sampled signals specified by a finite number of signal samples that represent continuous signals can be sharply bounded both in signal and spectral domains. This is quite obvious for some signal spectral representations such as Haar transform signal spectra. In particular, Haar basis functions are examples of sampled functions sharply bounded both in signal and Haar spectral domains. Another example represents Radon Transform: if object slice is limited in its extent, its projections are obviously also limited in the extent. Is this relevant also to DFT and DCT, that originate from the integral Fourier transform, which features the above uncertainty principle?

The answer is yes, it is. Such space-frequency sharply bounded signals can be generated using the above mentioned iterative Gershberg-Papoulis type algorithm that, at each iteration, applies required corresponding bounds alternatively in signal and spectral domains. An example of such a space-limited/band limited image is shown in Figure 15. Such images are very useful as test images for testing image processing algorithms.

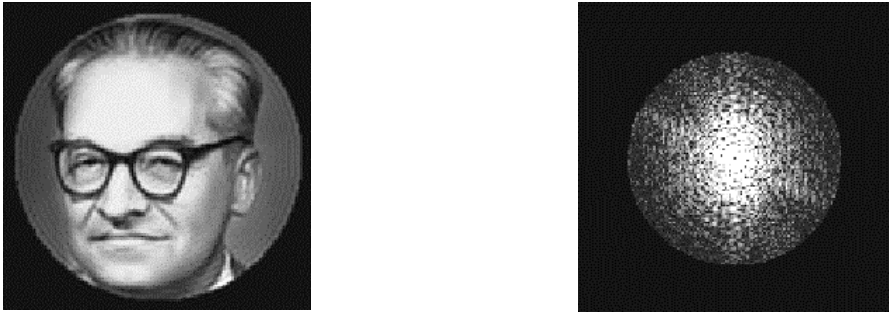

Figure 15. Space limited image "V. A. Kotelnikov" (the author of Ref. [ 42], left) and its band-limited DFT spectrum (right) shown centered at the signal dc component with spectrum intensity shown by image gray level.

In a similar way one can generate space-limited and band-limited analogs of the discrete sinc-function, functions, which form band-limited shift bases in the given space bounds. In Refs. [ 127], [131] such functions were coined a name "*sinc-lets*". Figure 16 shows an example of a sinc-let and its DFT spectrum in its three positions within a bounded interval and, for comparison, the discrete sinc-function for the same spectral band limitation.

The relationship between bounds in signal and DFT domains, i.e. between the number of signal non-zero samples $N_{sign}$ from $N$ samples of the signal sampling lattice and the number of signal non-zero spectral samples $N_{spect}$ is defined by the *discrete uncertainty principle*:

$$N_{sign} \times N_{spect} \geq N ,  \qquad (6)$$

which can be derived from the above continuous uncertainty principle ([ 127]) using known relationships between signal support, its bandwidth and sampling intervals in signal and spectral domains.

## 6. Conclusion

Two key properties of fast transforms that make them indispensable tool in digital imaging for image discretization, compression, restoration and enhancement are energy compaction capability and fast algorithms for transform computation. The multi resolution capability of wavelets adds substantially to these properties. Discrete Fourier Transform and Discrete Cosine Transform, being complementing each other discrete representations of the integral Fourier Transform, enable, in addition, fast digital convolution and are ideally suited for fast and accurate image resampling. In most applications, DCT is preferable to DFT thanks to its substantially lower vulnerability to boundary effects in digital filtering.

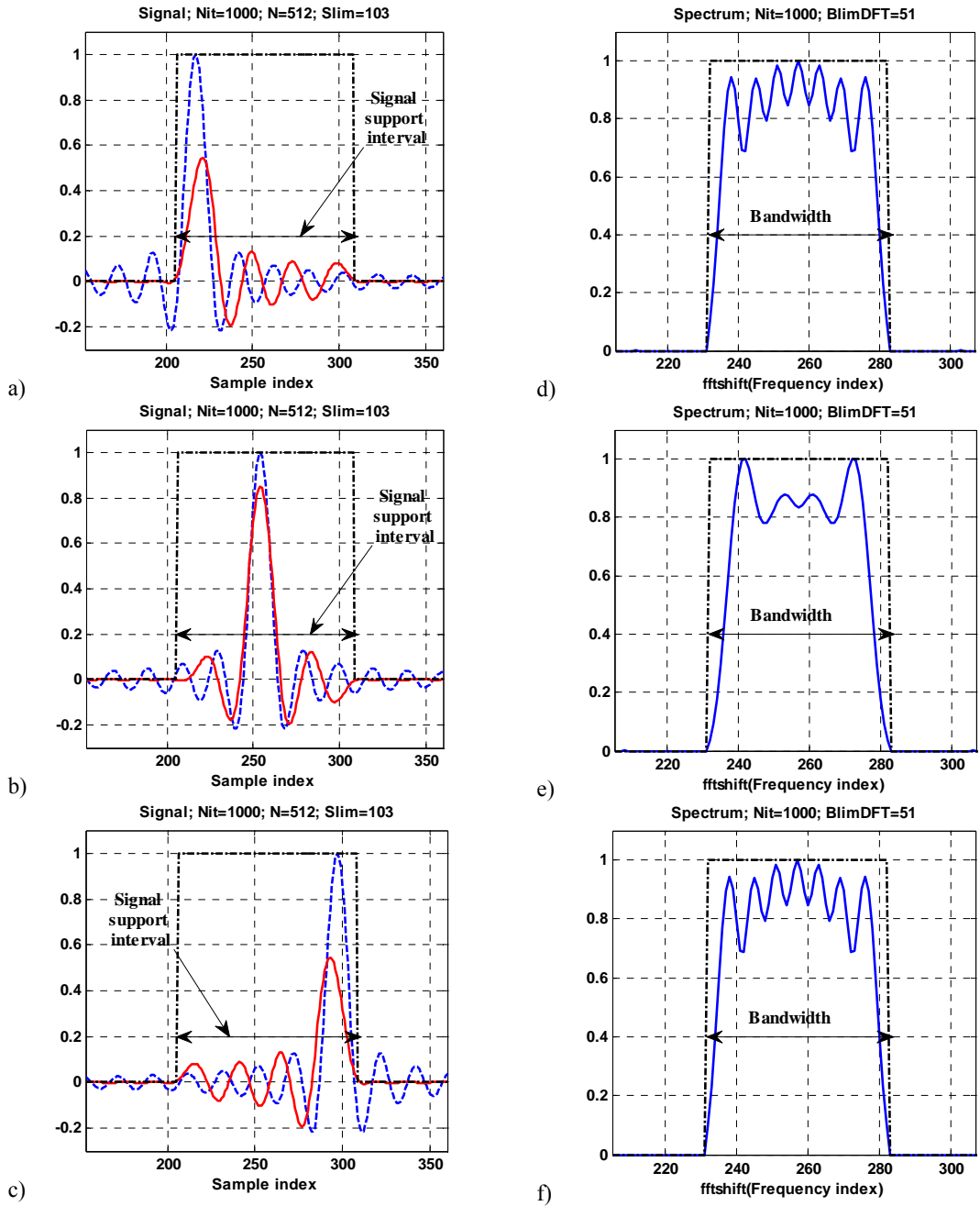

Figure 16. Examples of a "sinc-let" (red plots) and, for comparison, of the discrete sinc-function for the same band limitation (blue plots) in their three different positions within the interval of signal support of 103 samples of 512 samples (boxes a-c). Plots of their DFT spectra with bandwidth of 51 samples are shown in boxes (d-f).

High efficiency of such transforms as DCT and appropriately designed wavelets in compact representation of images evidences that they are in a good correspondence with statistics of real images. It is only natural to hypothesize that similar mechanisms may have evolved in visual systems of animals and humans. There are quite many indications in favor of such a hypothesis. For instance, basis functions of local DCT in window of 3x3 pixels resemble very much directional spatial receptive fields found in vertebrates ([132, 133]); similar resemblance was found between orthogonal cubic B-spline wavelet and receptive field of simple cortical cell ([134]). This interesting issue is a subject of many publications. Good reviews one can find in [135, 136]. Moreover, these analogies stimulate extending the linear transform decomposition of images toward nonlinear representations and efficient coding transforms that further increase the independence between image decomposition components and, therefore, transform energy compaction capability ([137-141]).

The major tendency in the imaging engineering nowadays is computational imaging. Computer processing of sensor data enables substantial price reduction and sometimes even complete removal of imaging optics and similar imaging hardware. It also gives birth to numerous new imaging techniques in astrophysics, biology, industrial engineering, remote sensing and other applications. No doubts, this area promises many new achievements in the coming years. And it is certain that fast transforms reviewed in the paper will remain to be the major tool in this process.

# Appendix

Table 2. Selected fast transforms and their computational complexity

| Transform name | | Transform definition in denotations: $a_{k,l}$ - image samples; $Sp_{r,s}$ - image transform coefficients; $\Delta x$, $\Delta f$ - sampling intervals in image and transform domains, correspondingly | Computational complexity (operations per sample; $C_{(.)}$ - various constants) | |
|---|---|---|---|---|
| | | | "Global": applied to entire signal of $N$ samples | "Local": applied in moving window of $WSz$ pixels |
| Discrete Fourier Transforms | Canonic DFT | $Sp(r) = \frac{1}{\sqrt{N}} \sum_{k=0}^{N-1} a_k \exp\left(i2\pi \frac{kr}{N}\right)$; 2D transform is separable to two 1D transforms | $C_{DFT}^G \log N$ | $C_{DFT}^L (\leq WSz)$ |
| | General Shifted Scaled DFT | $Sp(r) = \frac{1}{\sqrt{N}} \sum_{k=0}^{N-1} a_k \exp\left[i2\pi \frac{(k+u)(r+v)}{\sigma N}\right]$; 2D transform is separable to two 1D transforms. $u, v$ - shift parameters that indicate shifts $u\Delta x, v\Delta f$ of sampling lattices in signal and spectrum domains; $\sigma$ - sampling scale parameter ($\Delta x \Delta f = 1/\sigma N$) | $C_{SDFT}^G \log N$ | |
| | 2D Shifted Scaled Rotated DFT (ScRDFT) | $Sp(r,s) = \frac{1}{\sigma N} \sum_{k=0}^{N-1} \sum_{l=0}^{N-1} a_{k,l}$ $\exp\left[i2\pi \left(\frac{\tilde{k} \cos\theta + \tilde{l} \sin\theta}{\sigma N} \tilde{r} - \frac{\tilde{k} \sin\theta - \tilde{l} \cos\theta}{\sigma N} \tilde{s}\right)\right]$ $\tilde{k} = k + u_1; \tilde{r} = r + v_1; \tilde{l} = l + u_2; \tilde{s} = s + v_2$ $\theta$ - rotation angle | $C_{ScRDFT}^G \log N$ | n/a |
| Discrete Cosine Transform (DFT) | | $Sp(r) = \frac{1}{\sqrt{N}} \sum_{k=0}^{N-1} a_k \cos\left(2\pi \frac{k+1/2}{N} r\right)$; 2D transform is separable to two 1D transforms | $C_{DCT}^G \log N$ | $C_{DCT}^L (\leq WSz)$ |
| Walsh-Hadamard Transform | Hadamard Transform | $Sp(r) = \frac{1}{\sqrt{N}} \sum_{k=0}^{N-1} a_k had_k(r)$; $had_k(r) = (-1)^{\sum_{m=0}^{n-1} k_m r_m}$ $k = \sum_{m=0}^{n-1} k_m 2^m; r = \sum_{m=0}^{n-1} r_m 2^m; N = 2^n$ | Addition operations only $C_{WHT}^G \log N$ | n/a |
| | Walsh Transform | $Sp(r) = \frac{1}{\sqrt{N}} \sum_{k=0}^{N-1} a_k (-1)^{\sum_{m=0}^{n-1} k_m r_{n-m-1}^{GrayCode}}$; $r_{n-m-1}^{GrayCode}$ - binary digits of Gray code of $r$ taken in bit reversal order | | |
| Haar Transform (the simplest discrete wavelet transform) | | $Sp(r) = \frac{1}{\sqrt{N}} \sum_{k=0}^{N-1} a_k har_k(r)$; $har_k(r) = 2^{msb}(-1)^{k_{n-1-msb}} \delta(r_{\mod 2^{msb}} - \lfloor 2^{1-msb} k \rfloor)$ $msb$ - index of the most significant non-zero bit of binary code of $k$; $\lfloor . \rfloor$ - residual of $(.)$ | Addition operations only $C_{Haar}^G$ | n/a |

# Index